\documentclass{article}

\usepackage[nonatbib]{neurips_2020}
\usepackage[utf8]{inputenc} 
\usepackage[T1]{fontenc}    
\usepackage{hyperref}       
\usepackage{url}            
\usepackage{booktabs}       
\usepackage{amsfonts}       
\usepackage{nicefrac}       
\usepackage{microtype}      
\usepackage{graphicx}
\usepackage{caption}
\usepackage{makecell}
\usepackage[toc,page]{appendix}
\usepackage[multiple]{footmisc}

\title{TweetDrought: A Deep-Learning Drought Impacts Recognizer based on Twitter Data}

\begin{document}

\maketitle

\begin{abstract}

Acquiring a better understanding of drought impacts becomes increasingly vital under a warming climate. Traditional drought indices describe mainly biophysical variables and not impacts on social, economic, and environmental systems. We utilized natural language processing and bidirectional encoder representation from Transformers (BERT) based transfer learning to fine-tune the model on the data from the news-based Drought Impact Report (DIR) and then apply it to recognize seven types of drought impacts based on the filtered Twitter data from the United States. Our model achieved a satisfying macro-F1 score of 0.89 on the DIR test set. The model was then applied to California tweets and validated with keyword-based labels. The macro-F1 score was 0.58. However, due to the limitation of keywords, we also spot-checked tweets with controversial labels. 83.5\% of BERT labels were correct compared to the keyword labels. Overall, the fine-tuned BERT-based recognizer provided proper predictions and valuable information on drought impacts. The interpretation and analysis of the model were consistent with experiential domain expertise.
\end{abstract}

\section{Introduction}

Drought is a major but normal natural hazard with complex and broad impacts on many sectors of society and the economy \cite{hayes2012drought}. One significant characteristic of drought is the difficulty of determining boundaries in space and time, in contrast to other disasters such as floods and wildfires \cite{wilhite1985understanding}. Drought impacts range across wildlife and agriculture to public health and social stability. Climate change is impacting precipitation patterns on a large scale, which increases the frequency and intensity of extremes, including severe drought \cite{van2006impacts, field2012managing, moustakis2021seasonality}. Drought vulnerability and impacts are also partly the result of human activity \cite{aghakouchak2021anthropogenic,van2016drought,cook2018climate}. Improving insight into drought impacts can strengthen societal resilience to more severe droughts under a warming climate.

Many studies have employed remote sensing, meteorological and hydrological variables, crop yield, and other biomass data to develop drought indices to monitor and evaluate drought intensity, frequency, spatial and temporal extent. Some of them tried to link drought indices to drought impacts on agriculture and the ecohydrological environment \cite{tadesse2015assessing,stagge2015modeling,bachmair2017developing, de2020near}. However, a primary challenge of similar studies is that the input data sets for calculating indices fall short of capturing the diversity of drought impacts in social and economic sectors such as water supply and small business. Therefore, most current drought indices would underestimate the severity and scope of drought impacts in the human dimension due to the limitation of data.

 A 2020 study suggests that social media could be a good source of informative data to provide the different types of drought impacts, which could reflect the sensory information from people's observations and experiences in a water-short area \cite{smith2020calibrating}. Twitter, as one of the largest international social media platforms, has been increasingly applied to study natural hazards. For instance, a classic study employed tweets as its only input data to detect earthquakes' location in Japan based on probabilistic models \cite{sakaki2010earthquake}. Tweets were also used to extract the damages of disasters by applying various machine-learning methods, such as support vector machine, decision tree, and logistic regression \cite{ashktorab2014tweedr}. Another study compared different neural networks built on top of various word embedding models of Tweets in Hindi on detecting situational information during various disasters \cite{madichetty2020detection}. However, to the best of our knowledge, no study has tried to apply social media data and advanced text-mining and natural language processing (NLP) techniques to identify and analyze drought impacts.
 
This paper presents a deep-learning-based transfer learning framework to identify multi-category drought impacts from Twitter data in the United States. The Drought Impact Reporter (DIR) data with multi-labels was employed to fine-tune the pre-trained bidirectional encoder representation from Transformers (BERT). The fine-tuned model was then applied to the filtered tweets from 2017 to 2020 to identify drought impacts. To evaluate the model's performance on tweets with transfer learning, we labeled a group of tweets in California based on the keywords related to the different types of drought impacts as a calibration data set. The exploratory results from the recognizer reveal potential for learning about various drought impacts and their relationships and trends from tweets, and that would help us be prepared for drought in the warming climate.

\section{Data and Methods}

\subsection{Data}

We acquired 14,178 records of labeled DIR data from 2011 to 2020 in the United States from the National Drought Mitigation Center (NDMC) to fine-tune the pre-trained BERT model. The DIR collects drought impacts primarily from news media. The drought impacts are manually classified into nine categories \footnote[1]{The nine categories are agriculture, energy, plants \& wildlife, society \& public health, water supply \& quality, business \& industry, fire, relief, response \& restrictions, and tourism \& recreation} by experts at the NDMC. Considering the relationships of the different classes, we aggregated energy, business \& industry, and tourism \& recreation into a new category named economy to reduce the effects of imbalanced labels. Because a single drought event could affect multiple aspects in the natural environment, society, and economy, the learning objective of the model is a multi-label text classification problem. 

The Twitter data came from a previous research project at the NDMC \cite{smith2020calibrating}. The tweets from 2017 to 2020 in the United States were collected using the Twitter Archiving Google Spreadsheet (TAGS). Drought-related hashtags, such as \#drought and \#cadrought, were used to filter the drought-related tweets. Tweets were geo-located based on what users supplied as their location. 26,654 records of drought-related tweets were acquired among all states in the United States. Additional data descriptions are included in Appendix \ref{appendix:Description}.

\subsection{The Model Framework}

The multi-label recognizer for drought impacts was developed based on natural language processing (NLP) and the pre-trained BERT model. Our work was written using Python 3.7, and the primary packages are PyTorch, Hugging Face, and scikit-learn. The procedure is summarized into the following steps:

\textbf{Data preparation.} The text data preprocessing includes the following steps in order: i) removing HTML tags, URLs, and accented characters; ii) expanding contractions; and iii) removing special characters. The stop-words and numbers were left to preserve most of the information. The DIR records were randomly split into training, validation, and test set in 80\%, 10\%, and 10\%, respectively.

\textbf{Fine-tune the pre-trained BERT model.} We selected BERT as the primary model for the learning problem because it is a state-of-the-art model with outstanding performance in many NLP tasks such as general language understanding evaluation \cite{gonzalez2020comparing}. Besides, a fine-tuned BERT model will also achieve satisfying results on task-specific text classification \cite{devlin2018bert, sun2019fine}. Considering the size of the DIR data and the study's primary objective, we utilized the pre-trained BERT base (uncased) model with adding a dense layer and an output layer \cite{devlin2018bert, zhang2020dive}. The architecture of the fine-tuned BERT model is in Appendix \ref{appendix:BERT}. 

\textbf{Apply fine-tuned BERT model on tweets to identify drought impacts.} After fine-tuning the BERT model, the best-performing model on the DIR data set was saved to apply to drought-related tweets. Because the maximum length of tweets (43) was closed to the processed DIR records (44), the tokenized tweets could be directly inputted to the fine-tuned BERT model without truncating.

\textbf{Validate and analyze the predictions.} Because over 30\% of tweets (9,419) were located in California, we used them as a case study to validate the performance of the fine-tuned BERT model on tweets data. Keywords for each type of drought impact were generated from the top-100 unigrams based on the DIR data. If a tweet contained any of the keywords in a specific type of drought impact, it would be labeled as one for the kind. A table of keywords is in Appendix \ref{appendix:keywords}. In the end, 1817 tweets in California were labeled and considered as true samples to validate the results from the model. Although this validation process was very efficient, we realized that the limitation of the keywords might lead to potential biases. Therefore, additional manual spot-check tests were used to examine the validations and analyze the predictions.

\section{Results and Discussion}

\subsection{The Fine-tuned BERT on DIR}

\begin{table}[htbp]
  \caption{Summary of the fine-tuned BERT's performance on the DIR test data set.}
  \label{tab:DIR}
  \centering
  \begin{tabular}{lccc}
    \toprule
    Category of Drought Impacts & Recall & Precision & F1\\
    \midrule
    Overall (micro/macro) & 0.86/0.82 & 0.95/0.95  & 0.90/0.87 \\
    \midrule
    Agriculture & 0.93 & 0.98  & 0.96    \\
    Economy & 0.72 & 0.95  & 0.85    \\
    Fire & 0.88 & 0.97 & 0.92 \\
    Plants \& Wildlife & 0.78 & 0.88 & 0.83 \\
    Relief, Response \& Restrictions & 0.92 & 0.93 & 0.93 \\
    Society \& Public Health & 0.56 & 0.98 & 0.72 \\
    Water Supply \& Quality & 0.87  & 0.92  & 0.89 \\
    \bottomrule
  \end{tabular}
\end{table}

To evaluate the performance of the fine-tuned BERT, we calculated the precision, recall, and F1 score on the test data set. Table \ref{tab:DIR} is a summary of the metrics. Except for the impact on society and public health, the model's F1 scores on the rest types range from 0.83 to 0.96. Although the recall on society and public health is the lowest (0.56), the model achieved the highest precision on this type of impact (0.98). Based on our empirical knowledge, it is reasonable that the model had lower sensitivity and performance on identifying the drought impacts on socio-economic sectors. These types of impacts are extensive and particularly heterogeneous within the category. And as compared to agriculture and fire, fewer impacts were observed and recorded in the DIR. The micro-F1 on the test set is 0.90 and slightly higher than the unweighted mean (macro-) F1 (0.87). A higher recall is better for identifying drought impacts than precision to attain better mitigation. Thus, we assigned priority to recall in assessing models. The macro-recall on the DIR test set is 0.82. Overall, the BERT model was successfully fine-tuned on the DIR data to identify various drought impacts. Confusion matrices on the DIR data are included in Appendix \ref{appendix:cm_dir}.

\subsection{Transfer Learning on Tweets}

The fine-tuned BERT model was then applied to the drought-related tweets in the U.S. to identify the types of impacts. And the keyword-based labels in California were used to validate the model predictions. Table \ref{tab:Tweets} presents the metrics based on the keyword labels. The macro-F1 score is 0.58, and the macro-recall is 0.67. The BERT model had the best performance on identifying fire impacts and the worst on society and public health. If we exclude the worst-performing label, the macro-F1 and recall improve to 0.65 and 0.71. Confusion matrices are included in Appendix \ref{appendix:cm_tweets}. Since the keywords of the fire were straightforward and distinctive, such as \emph{wildfires} and \emph{burn}, the drought impacts on fire could be explicitly described. However, other impacts, such as society and public health, are not well-defined. The distinctions between the impacts are hard to identify by keywords. For example, words to describe the impacts might be interchangeable, and a phrase might indicate multiple impacts. Hence, the characteristics of drought impacts and complexity of natural language strain the limits of keyword-based labels as ground truth. 

\begin{table}[htbp]
  \caption{Summary of the fine-tuned BERT's performance on keyword-labeled tweets in CA.}
  \label{tab:Tweets}
  \centering
  \begin{tabular}{lccc}
    \toprule
    Category of Drought Impacts & Recall & Precision & F1\\
    \midrule
    Overall (micro/macro) & 0.72/0.67 & 0.52/0.58 & 0.60/0.58 \\
    \midrule
    Agriculture & 0.54 & 0.78 & 0.63    \\
    Economy & 0.42 & 0.44  & 0.43    \\
    Fire & 0.81 & 0.95 & 0.87 \\
    Plants \& Wildlife & 0.65 & 0.67 & 0.66 \\
    Relief, Response \& Restrictions & 0.81 & 0.52 & 0.63 \\
    Society \& Public Health & 0.58 & 0.09 & 0.15 \\
    Water Supply \& Quality & 0.92  & 0.59  & 0.72 \\
    \bottomrule
  \end{tabular}
\end{table}

To have a more reliable evaluation, we manually spot-checked some tweets for each label in California, comparing the differences between BERT and keyword labels. Due to the paper's length, we primarily discuss the impacts on agriculture and on society and public health. Agriculture is the most apparent impact of drought. And the BERT model had the lowest performance on recognizing the drought impacts on society and public health.

\textbf{Agriculture:} We randomly examined 100 tweets with controversial labels, and 89 were more rational with BERT labels. In 39 tweets with false-positive (FP) labels, 36 were related to food and soil moisture, which are however implicitly connected to agriculture. The model BERT is likely to have a better generalization capability and can identify impacts from sentences rather than words. In 61 tweets with false-negative (FN) labels, 53 were labeled as impacts on plants and wildlife by BERT, which had contents about \emph{parks, trees, lawn, grass,} and \emph{irrigation}. Compared to keyword-based labels, the BERT model successfully distinguished between agricultural and urban impacts, especially related to irrigation.

\textbf{Society \& Public Health} 100 tweets with controversial labels on society and public health were also randomly checked, and 78 could be explained appropriately with BERT labels. Compared to agriculture, drought impacts on society and public health are more abstract to describe, which is a primary reason for the worse model performance. In 64 tweets with FP labels, 20 reflected personal feelings about drought, such as worried, frustrated, and hopeful. 50 out of 64 tweets were also labeled with the drought impacts on water supply and quality. Therefore, these tweets could be related to society and public health if we consider a more general definition. 22 out of the 45 FN tweets were labeled with agriculture because their contents included food security and crops. The vague concept of drought impacts in the human dimension makes it challenging for the BERT model, even for domain experts, to categorize impacts.

Moreover, we investigate the interconnection between each type of impact to verify if the BERT labels could reveal the common sense of drought impacts. The analysis was applied to the predicted labels in the whole country. The most common associated label with agriculture impacts is water supply and quality (24\%). The impacts on fire are mostly combined with the impacts on society and public health (26\%). Impressively, 78\% drought impacts on water supply and quality happen with impacts on relief, response, and restrictions. The relationships between drought impacts identified by the BERT model perfectly match our empirical experience.

\section{Conclusion and Future Work}

With the increasing intensity and frequency of drought under the changing climate, it becomes crucial to acquire a deeper insight into drought impacts. However, traditional drought indices using hydro-meteorological and remote sensing variables tend to underestimate the drought impacts in the human dimension. This paper proposed a framework using the state-of-the-art BERT model and transfer learning to recognize the types of impacts in drought-related tweets from the United States. The pre-trained BERT model was utilized and fine-tuned on the DIR data multi-labeled in seven different drought impact categories. The macro-F1 score and recall on the DIR test set are 0.87 and 0.82. The fine-tuned BERT model was then applied to the drought-related tweets from 2017 to 2020. To evaluate the model's performance, we built a list of keywords for each type of drought impact. The macro-F1 score and recall on keyword-labeled tweets are 0.58 and 0.67. We also manually spot-checked tweets in each type of impact and compared the labels between BERT and keywords. Drought impacts on agriculture and society and public health were particularly analyzed in the paper to investigate the performance of the fine-tuned BERT. For the agriculture impacts, 89\% of the controversial tweets (FP, FN) were more rational with BERT-based labels. For society and public health, the proportion was 78\%. Compared to keywords, the fine-tuned BERT had a better generalization capability and sensitivity to drought impacts. Also, the BERT model distinguished between drought impacts in rural and urban areas. Overall, the BERT-based recognizer provided promising predictions on the types of drought impacts. From interpreting the results from the BERT on drought-related tweets, we find an efficient way to recognize drought impacts in the natural environment and human dimension.

Further studies are also recommended to analyze and interpret the BERT predictions of drought impacts. For example, it is worth mapping the spatial patterns of the various drought impacts to study the trends in different states and climate regimes. Although the filtered tweets were related to drought, many were general thoughts rather than descriptions of specific impacts. Therefore, another NLP model could be developed to identify whether tweets include impact information rather than just a demonstration of drought awareness. This means of tapping into what people are saying about their drought experience can provide better data to prepare for and respond to drought and climate change.

\newpage
\bibliographystyle{IEEEtran}
\bibliography{reference.bib}

\begin{thebibliography}{10}
\providecommand{\url}[1]{#1}
\csname url@samestyle\endcsname
\providecommand{\newblock}{\relax}
\providecommand{\bibinfo}[2]{#2}
\providecommand{\BIBentrySTDinterwordspacing}{\spaceskip=0pt\relax}
\providecommand{\BIBentryALTinterwordstretchfactor}{4}
\providecommand{\BIBentryALTinterwordspacing}{\spaceskip=\fontdimen2\font plus
\BIBentryALTinterwordstretchfactor\fontdimen3\font minus
  \fontdimen4\font\relax}
\providecommand{\BIBforeignlanguage}[2]{{%
\expandafter\ifx\csname l@#1\endcsname\relax
\typeout{** WARNING: IEEEtran.bst: No hyphenation pattern has been}%
\typeout{** loaded for the language `#1'. Using the pattern for}%
\typeout{** the default language instead.}%
\else
\language=\csname l@#1\endcsname
\fi
#2}}
\providecommand{\BIBdecl}{\relax}
\BIBdecl

\bibitem{hayes2012drought}
M.~J. Hayes, M.~D. Svoboda, B.~D. Wardlow, M.~C. Anderson, and F.~Kogan,
  ``Drought monitoring: Historical and current perspectives,'' 2012.

\bibitem{wilhite1985understanding}
D.~A. Wilhite and M.~H. Glantz, ``Understanding: the drought phenomenon: the
  role of definitions,'' \emph{Water international}, vol.~10, no.~3, pp.
  111--120, 1985.

\bibitem{van2006impacts}
M.~K. Van~Aalst, ``The impacts of climate change on the risk of natural
  disasters,'' \emph{Disasters}, vol.~30, no.~1, pp. 5--18, 2006.

\bibitem{field2012managing}
C.~B. Field, V.~Barros, T.~F. Stocker, and Q.~Dahe, \emph{Managing the risks of
  extreme events and disasters to advance climate change adaptation: special
  report of the intergovernmental panel on climate change}.\hskip 1em plus
  0.5em minus 0.4em\relax Cambridge University Press, 2012.

\bibitem{moustakis2021seasonality}
Y.~Moustakis, S.~M. Papalexiou, C.~J. Onof, and A.~Paschalis, ``Seasonality,
  intensity, and duration of rainfall extremes change in a warmer climate,''
  \emph{Earth's Future}, vol.~9, no.~3, 2021.

\bibitem{aghakouchak2021anthropogenic}
A.~AghaKouchak, A.~Mirchi, K.~Madani, G.~Di~Baldassarre, A.~Nazemi, A.~Alborzi,
  H.~Anjileli, M.~Azarderakhsh, F.~Chiang, E.~Hassanzadeh \emph{et~al.},
  ``Anthropogenic drought: Definition, challenges, and opportunities,'' 2021.

\bibitem{van2016drought}
A.~F. Van~Loon, T.~Gleeson, J.~Clark, A.~I. Van~Dijk, K.~Stahl, J.~Hannaford,
  G.~Di~Baldassarre, A.~J. Teuling, L.~M. Tallaksen, R.~Uijlenhoet
  \emph{et~al.}, ``Drought in the anthropocene,'' \emph{Nature Geoscience},
  vol.~9, no.~2, pp. 89--91, 2016.

\bibitem{cook2018climate}
B.~I. Cook, J.~S. Mankin, and K.~J. Anchukaitis, ``Climate change and drought:
  From past to future,'' \emph{Current Climate Change Reports}, vol.~4, no.~2,
  pp. 164--179, 2018.

\bibitem{tadesse2015assessing}
T.~Tadesse, B.~D. Wardlow, J.~F. Brown, M.~D. Svoboda, M.~J. Hayes, B.~Fuchs,
  and D.~Gutzmer, ``Assessing the vegetation condition impacts of the 2011
  drought across the us southern great plains using the vegetation drought
  response index (vegdri),'' \emph{Journal of Applied Meteorology and
  Climatology}, vol.~54, no.~1, pp. 153--169, 2015.

\bibitem{stagge2015modeling}
J.~H. Stagge, I.~Kohn, L.~M. Tallaksen, and K.~Stahl, ``Modeling drought impact
  occurrence based on meteorological drought indices in europe,'' \emph{Journal
  of Hydrology}, vol. 530, pp. 37--50, 2015.

\bibitem{bachmair2017developing}
S.~Bachmair, C.~Svensson, I.~Prosdocimi, J.~Hannaford, and K.~Stahl,
  ``Developing drought impact functions for drought risk management,''
  \emph{Natural Hazards and Earth System Sciences}, vol.~17, no.~11, pp.
  1947--1960, 2017.

\bibitem{de2020near}
M.~M. de~Brito, C.~Kuhlicke, and A.~Marx, ``Near-real-time drought impact
  assessment: A text mining approach on the 2018/19 drought in germany,''
  \emph{Environmental Research Letters}, 2020.

\bibitem{smith2020calibrating}
K.~H. Smith, A.~J. Tyre, Z.~Tang, M.~J. Hayes, and F.~A. Akyuz, ``Calibrating
  human attention as indicator monitoring\# drought in the twittersphere,''
  \emph{Bulletin of the American Meteorological Society}, vol. 101, no.~10, pp.
  E1801--E1819, 2020.

\bibitem{sakaki2010earthquake}
T.~Sakaki, M.~Okazaki, and Y.~Matsuo, ``Earthquake shakes twitter users:
  real-time event detection by social sensors,'' pp. 851--860, 2010.

\bibitem{ashktorab2014tweedr}
Z.~Ashktorab, C.~Brown, M.~Nandi, and A.~Culotta, ``Tweedr: Mining twitter to
  inform disaster response.'' pp. 269--272, 2014.

\bibitem{madichetty2020detection}
S.~Madichetty and S.~Muthukumarasamy, ``Detection of situational information
  from twitter during disaster using deep learning models,''
  \emph{S{\=a}dhan{\=a}}, vol.~45, no.~1, pp. 1--13, 2020.

\bibitem{gonzalez2020comparing}
S.~Gonz{\'a}lez-Carvajal and E.~C. Garrido-Merch{\'a}n, ``Comparing bert
  against traditional machine learning text classification,'' \emph{arXiv
  preprint arXiv:2005.13012}, 2020.

\bibitem{devlin2018bert}
J.~Devlin, M.-W. Chang, K.~Lee, and K.~Toutanova, ``Bert: Pre-training of deep
  bidirectional transformers for language understanding,'' \emph{arXiv preprint
  arXiv:1810.04805}, 2018.

\bibitem{sun2019fine}
C.~Sun, X.~Qiu, Y.~Xu, and X.~Huang, ``How to fine-tune bert for text
  classification?'' in \emph{China National Conference on Chinese Computational
  Linguistics}.\hskip 1em plus 0.5em minus 0.4em\relax Springer, 2019, pp.
  194--206.

\bibitem{zhang2020dive}
A.~Zhang, Z.~C. Lipton, M.~Li, and A.~J. Smola, \emph{Dive into Deep Learning},
  2020, \url{https://d2l.ai}.

\end{thebibliography}
\small
\nocite{*}

\newpage
\begin{appendices}

\section{Descriptive statistics for the DIR and tweets}
\label{appendix:Description}

\begin{figure}[htbp]
  \centering
  \includegraphics[width=\textwidth]{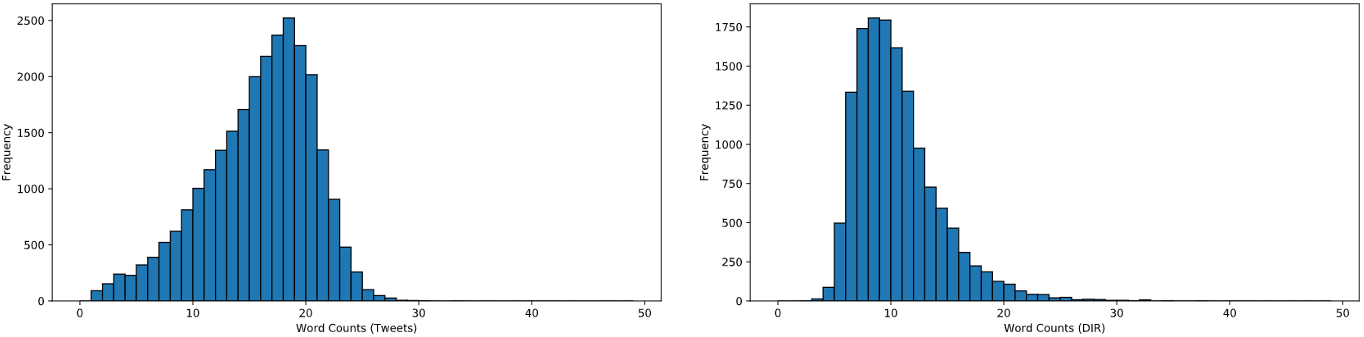}
  \caption{Histograms of word counts of the drought-related tweets (left) and the DIR data (right).}
  \label{fig:histograms}
\end{figure}

\begin{figure}[htbp]
  \centering
  \includegraphics[scale=0.2]{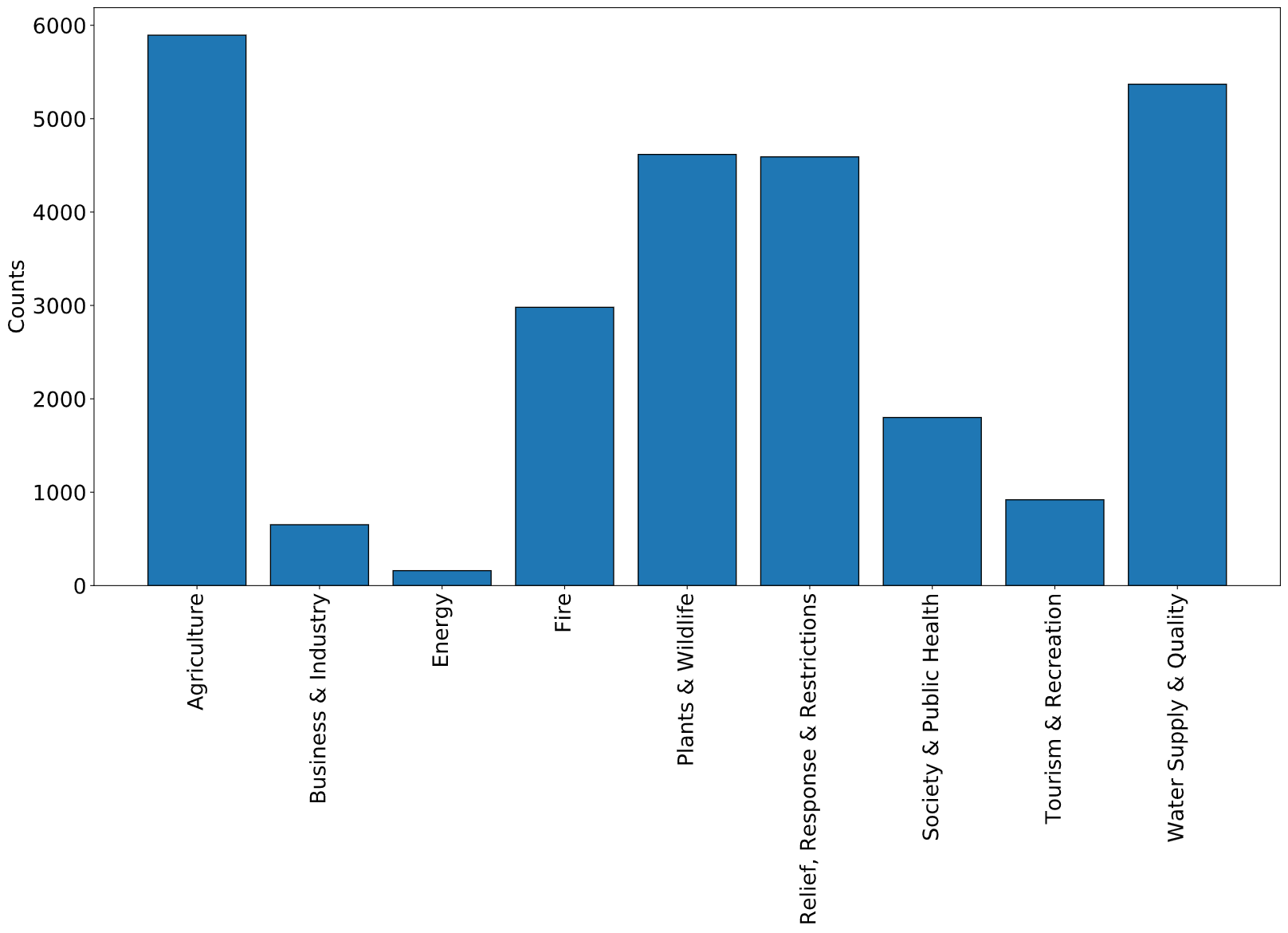}
  \caption{Label distribution of the DIR data.}
  \label{fig:label_distribution}
\end{figure}

The summary titles for each DIR record were used as input data to fine-tune the model because they had similar lengths to the tweets. Figure \ref{fig:histograms} are two histograms of word counts of the tweets and the title of the DIR records. The average length of tweets is about fifteen words. And the total number of words is 408,890. The average length of the title is about ten words. The total number of words is 143,260. Figure \ref{fig:label_distribution} is the distribution of the nine categories in the DIR data. Business \& industry, energy, tourism \& recreation, and society \& public health are minor groups. By employing the aggregated economy label, the effect of imbalanced label distribution was reduced. Agriculture and water supply \& quality are the top-two frequent impacts.

\section{The architecture of the fine-tuned BERT model}
\label{appendix:BERT}

\begin{figure}[htbp]
  \centering
  \includegraphics[width=\textwidth]{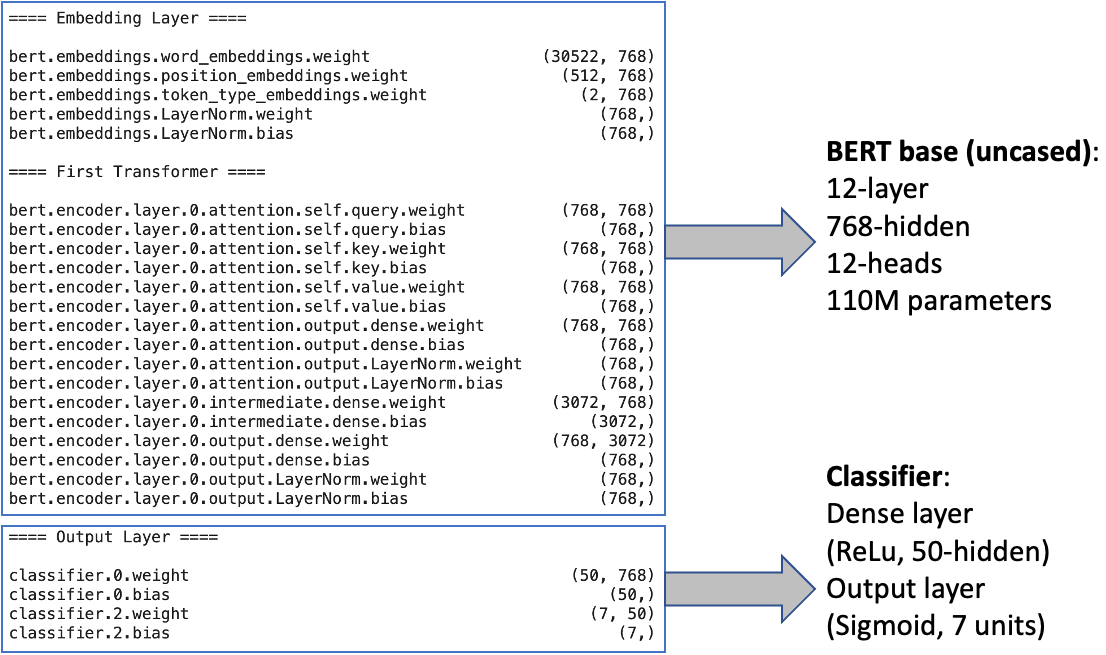}
  \caption{The architecture of the fine-tuned BERT model.}
  \label{fig:BERT}
\end{figure}

This study employed the BERT base uncased model pre-trained on English Wikipedia and BooksCorpus \cite{devlin2018bert}. The pre-trained BERT model has 12 layers, 768 hidden, and 12 heads. Figure \ref{fig:BERT} presents the architecture of the complete model. A 50-hidden dense layer with a ReLu activation function and a 7-unit output layer with a Sigmoid function were added to the BERT model to predict multi-labeled drought impacts. Since the output of the Sigmoid function is the probability, 0.5 was used as the threshold to convert the numbers to binary labels. The binary-cross-entropy loss function and Adam optimizer were applied to fine-tune the BERT.

\section{Keywords table}

Table \ref{tab:keywords} is a summary of the keywords that were applied to label the tweets in California as ground truth.

\label{appendix:keywords}
\begin{table}[htbp]
  \caption{Table of the keywords to label tweets in California}
  \label{tab:keywords}
  \centering
  \begin{tabular}{lc}
    \toprule
    Drought Impacts & Keywords\\
    \midrule
    Agriculture & \makecell[l]{hay, crops, corn, cattle, livestock, crop, farmers, wheat, pasture,\\irrigation, grass, producer, agriculture, grazing, cotton, yield, yields,\\soybean, pasture, ranchers, trees, producers, ponds, growth, growing}\\
    \midrule
    Economy & \makecell[l]{boat, ski, business, fishing, park, ramps, power, businesses, fireworks, \\golf, hydropower, lawn, prices}\\
    \midrule
    Fire & \makecell[l]{fire, burn, fires, wildfires, burning, burned, wildfire}\\
    \midrule
    Plants \& Wildlife & \makecell[l]{leaves, brown, plants, leaves, wildlife, fish, soil, lawn, garden, deer,\\browning, birds, bird, tree, trees}\\
    \midrule
    Relief, Response \& Restrictions & \makecell[l]{restrictions, ban, conservation, mandatory, voluntary, declaration,\\ governor, communities, prohibited, conserve}\\
    \midrule
    Society \& Public Health & \makecell[l]{quality, dust, food, health, allergies, smoke, homeowners,mental,stress}\\
    \midrule
    Water Supply \& Quality & \makecell[l]{restrictions, river, conservation, lake, irrigation, wells, ponds, quality,\\reservoir, pond}\\
    \bottomrule
  \end{tabular}
\end{table}

\section{Confusion metrics on the DIR test data set}
\label{appendix:cm_dir}

Figure \ref{appendix:cm_dir} shows confusion matrices of the fine-tuned BERT model on the DIR test data set.

\begin{figure}[htbp]
  \centering
  \includegraphics[width=\textwidth]{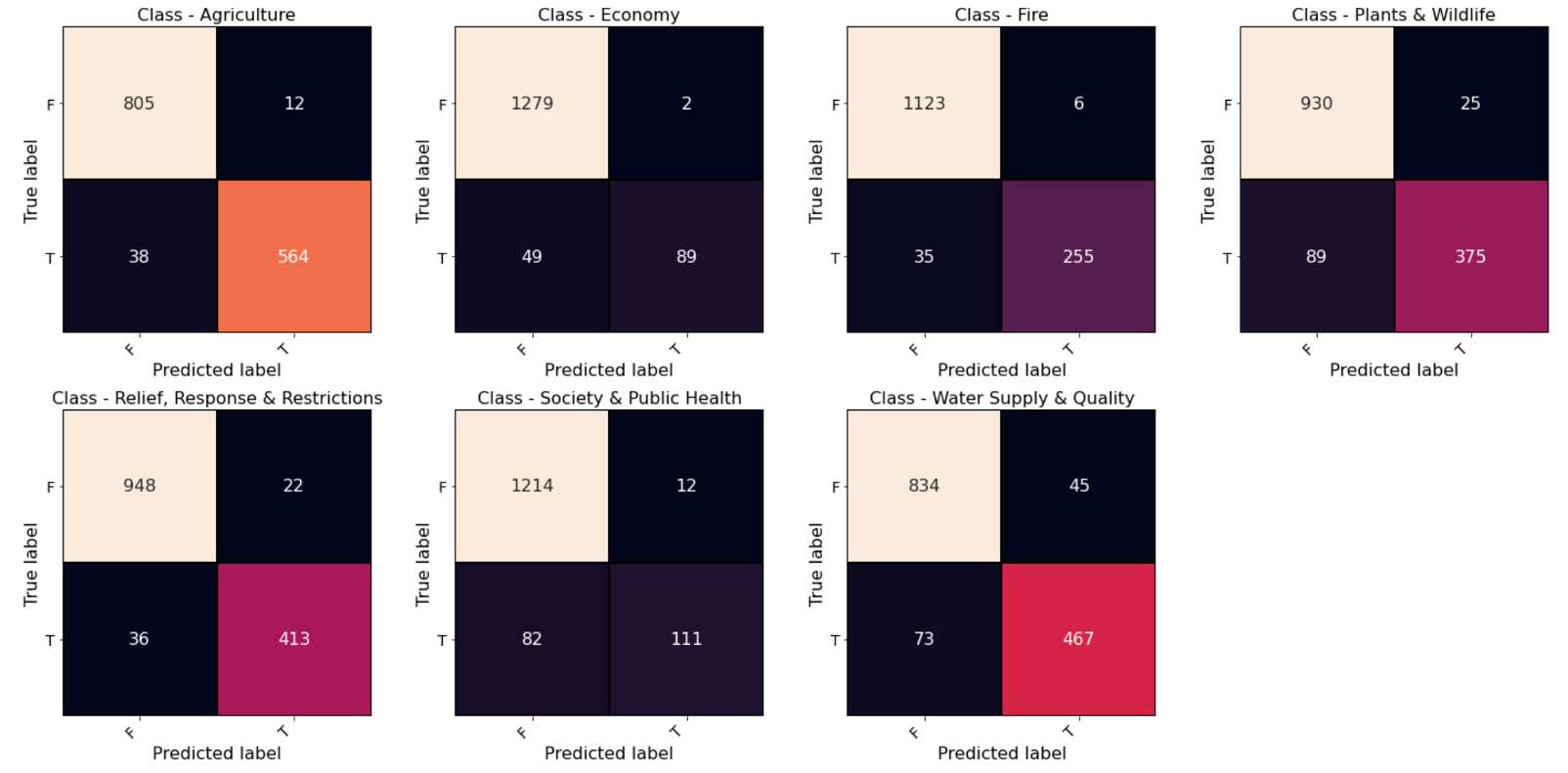}
  \caption{Confusion matrices of the fine-tuned BERT model on the DIR test data set.}
  \label{fig:cm_dir}
\end{figure}

\newpage
\section{Confusion metrics on the keyword-labeled tweets in California}
\label{appendix:cm_tweets}

Figure \ref{appendix:cm_tweets} shows confusion matrices of the fine-tuned BERT model on the keyword-labeled tweets in California.

\begin{figure}[htbp]
  \centering
  \includegraphics[width=\textwidth]{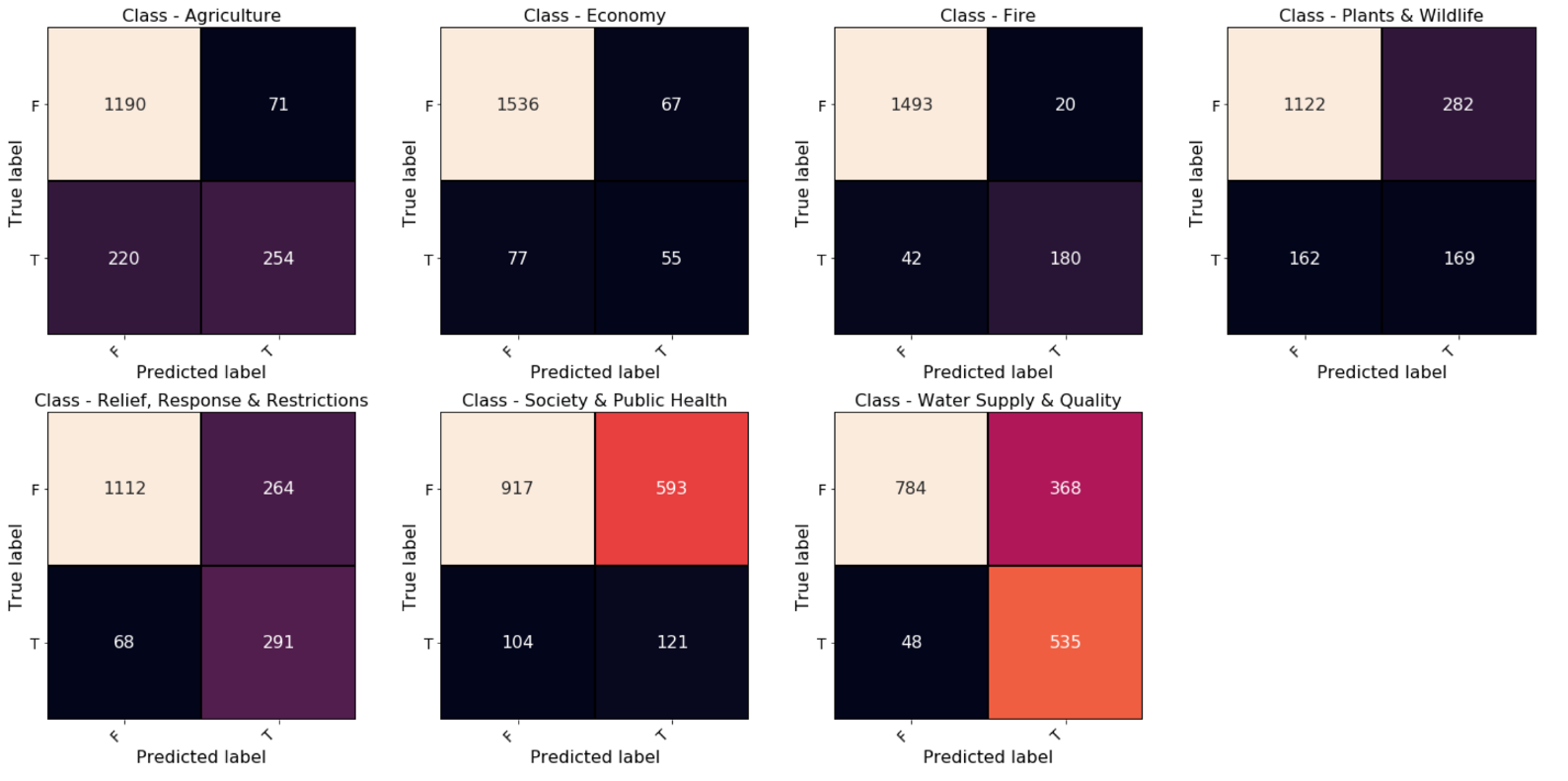}
  \caption{Confusion matrices of the fine-tuned BERT model on the keyword-labeled tweets in California.}
  \label{fig:cm_tweets}
\end{figure}

\end{appendices}

\end{document}